\documentclass[letterpaper, 10 pt, conference]{ieeeconf}  

\IEEEoverridecommandlockouts                              

\usepackage{cite}

\usepackage{amsmath,amssymb,amsfonts,amsthm}
\usepackage{algorithmic}
\usepackage{algorithm}
\usepackage{graphicx}
\usepackage{textcomp}
\usepackage{xcolor}
\usepackage{svg}
\usepackage{dsfont}
\usepackage{adjustbox}
\usepackage{url}

\usepackage{xspace}
\providecommand{\FullStop}{\text{~\@.\xspace}}
\providecommand{\Comma}{\text{~,\xspace}}

\title{\LARGE \bf
A Collision-Free Sway Damping Model Predictive Controller for Safe and Reactive Forestry Crane Navigation
}

\author{Marc-Philip Ecker$^{1,2}$, Christoph Fröhlich$^{2}$, Johannes Huemer$^{2}$, David Gruber$^{2}$,\\ Bernhard Bischof$^{2}$, Tobias Glück$^{2}$ and Wolfgang Kemmetmüller$^{1}$
\thanks{$^{1}$Marc-Philip Ecker and Wolfgang Kemmetmüller are with the
Automation \& Control Institute (ACIN), TU Wien, 1040 Vienna, Austria
        {\tt\small \{ecker,kemmetmueller\}@acin.tuwien.ac.at}}%
\thanks{$^{2}$Marc-Philip Ecker, Christoph Fröhlich, Johannes Huemer, David Gruber, Bernhard Bischof  and Tobias Glück are with the Center for Vision, Automation \& Control,
AIT Austrian Institute of Technology GmbH, 1210 Vienna, Austria
        {\tt\small \{marc-philip.ecker,christoph.froehlich,\newline johannes.huemer,david.gruber,bernhard.bischof,\newline tobias.glueck\}@ait.ac.at}}%
}

\begin{document}
\maketitle
\thispagestyle{empty}
\pagestyle{empty}

\begin{abstract}
Forestry cranes operate in dynamic, unstructured outdoor environments where simultaneous collision avoidance and payload sway control are critical for safe navigation. Existing approaches address these challenges separately, either focusing on sway damping with predefined collision-free paths or performing collision avoidance only at the global planning level. We present the first collision-free, sway-damping model predictive controller (MPC) for a forestry crane that unifies both objectives in a single control framework. Our approach integrates LiDAR-based environment mapping directly into the MPC using online Euclidean distance fields (EDF), enabling real-time environmental adaptation. The controller simultaneously enforces collision constraints while damping payload sway, allowing it to (i) replan upon quasi-static environmental changes, (ii) maintain collision-free operation under disturbances, and (iii) provide safe stopping when no bypass exists. Experimental validation on a real forestry crane demonstrates effective sway damping and successful obstacle avoidance. A video can be found at {\color{blue}\url{https://youtu.be/tEXDoeLLTxA}}.
\end{abstract}
\begin{figure}[h]
\centering
\includegraphics[trim=5cm 0.2cm 1.25cm 0cm,clip,scale=0.62]
{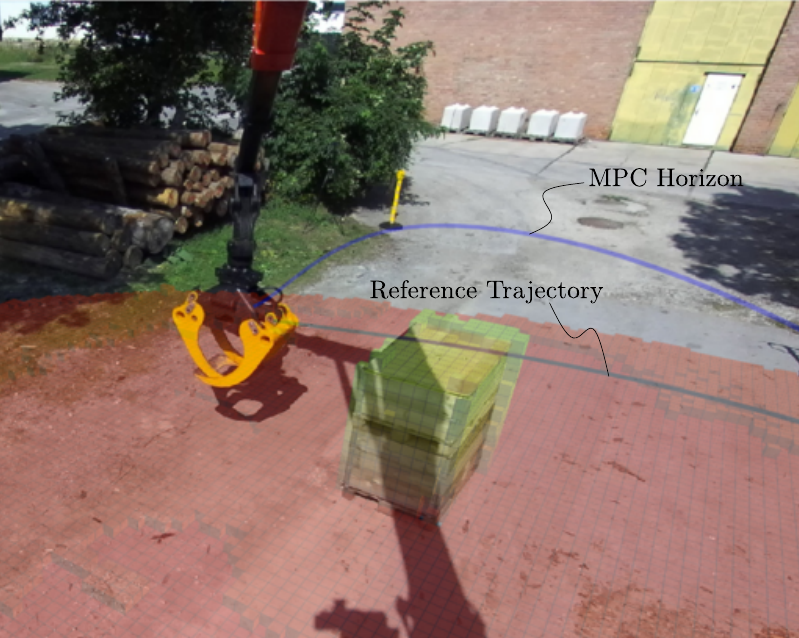}
\includegraphics[scale=0.25]{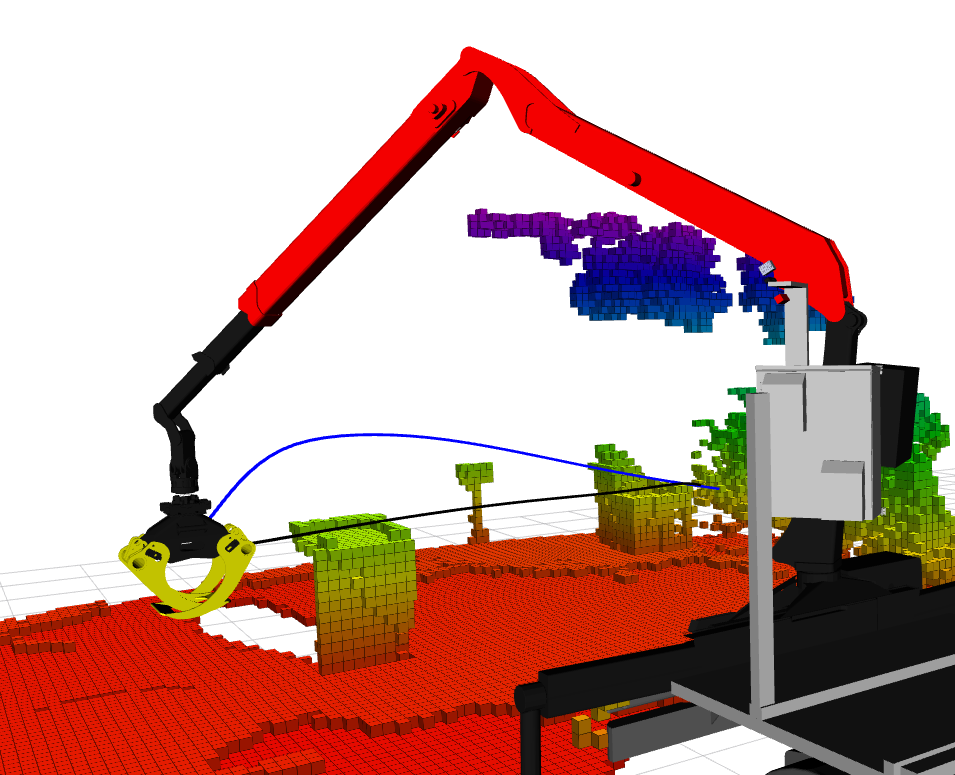}
\caption{Forestry crane equipped with LiDAR-based perception system. The proposed MPC framework enables autonomous navigation with simultaneous collision avoidance and sway damping capabilities for safe operation.}
\label{fig:KinematicChain}
\end{figure}

\section{Introduction}
Cranes are essential to modern forestry operations, handling and transporting timber throughout the harvesting and logistics chain. Safe operation demands exceptional skill, as operators must coordinate multiple hydraulic joints while navigating collision hazards in cluttered outdoor environments. An aging workforce combined with high training costs has created a critical shortage of qualified operators, driving urgent demand for autonomous and semi-autonomous control solutions to sustain both productivity and safety~\cite{agostini:2003,starr:2005,morales:2014,kalmari:2014,kalmari:2017,song:2020,dhakate:2022,ayoub:2024}.
Forestry cranes present fundamentally different challenges than industrial manipulators. These large-scale, underactuated systems operate in unstructured, dynamic environments where precise environmental knowledge is limited. Hydraulic actuation dynamics, payload sway, and perception uncertainties make accurate trajectory tracking particularly difficult, yet collision-free operation remains critical. Traditional tracking-based controllers struggle in these conditions, as tracking errors can lead directly to collisions.
To overcome these limitations, we propose a collision-free, sway-damping Model Predictive Controller (MPC) that integrates LiDAR-based perception directly into the control loop. Rather than relying on precise tracking of pre-computed collision-free paths, our approach embeds collision constraints directly within the MPC optimization, enabling real-time adaptation to environmental changes while simultaneously damping payload sway. This unified framework enhances safety and robustness by compensating for tracking inaccuracies and environmental uncertainties at the control level. We demonstrate the effectiveness of our approach through comprehensive real-world experiments on a forestry crane.

\begin{figure*}
\centering
\includegraphics[scale=0.48]{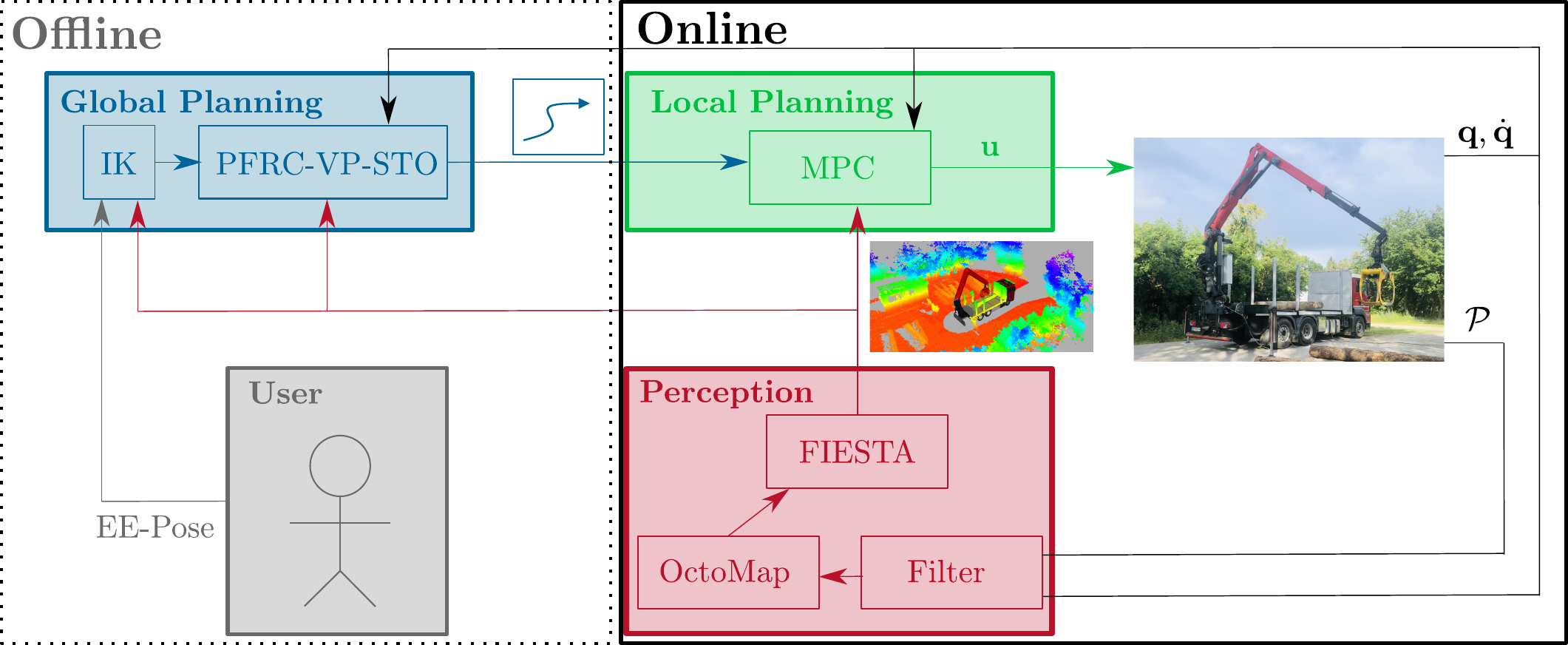}
\caption{Schematic of our navigation pipeline of the forestry crane. We use the global planner from \cite{ecker:2025} together with the EDF-based collision detection routine \cite{ecker:iros:2025}. To obtain a map, we apply a point cloud filter to remove the crane points, and use OctoMap \cite{hornung:2013} in combination with the FIESTA EDF mapping algorithm \cite{han:2019}. The perception-based collision-free sway damping MPC is the contribution of this work.}
\label{fig:Schematic}
\end{figure*}

\subsection{Related Work}

Automation of large-scale underactuated manipulators, such as the forestry crane depicted in Fig.~\ref{fig:KinematicChain}, has attracted increasing attention in recent years, see e.g. \cite{morales:2014,ayoub:2024,spinelli:2025}.
In \cite{kalmari:2014,kalmari:2017}, a model predictive path-tracking controller for sway damping of forestry cranes is introduced, demonstrating tracking performance on predefined paths. Contrary, a reinforcement learning (RL)-based end-effector tracking controller for a hydraulic manipulator is presented in \cite{dhakate:2022}, but sway damping is not addressed. Both approaches assume predefined collision-free paths with proper safety margin.
Jebellat et al. \cite{jebellat:2023,jebellat:2024} propose an open-loop, offline trajectory planning approach for sway damping based on dynamic programming. However, their evaluation is limited to simulations, without experiments on real machines. In addition, the planner is not able to handle collisions.
Ayoub et al. \cite{ayoub:2024} propose a fully autonomous log loading pipeline for a forestry crane, employing an RRT*-based local planner combined with a simple tracking controller that neither handles pendulum sway nor collision avoidance. Similarly, Spinelli et al. \cite{spinelli:2025} present a planning and control framework for large-scale underactuated material handling manipulators using LiDAR-based maps. In this approach, collision avoidance is handled by an RRT*-based global planner, while trajectory tracking is performed by a RL controller.
In \cite{ecker:2025}, a hybrid motion planning framework for forestry cranes is presented, combining a near time-optimal global planner with an MPC-based local planner. The global planner accounts for the hydraulic pump flow rate and significantly outperforms a classical RRT*-based approach. In \cite{ecker:ifac:2025}, the global planner is extended to account for the pendulum motion. However, these works emphasize theoretical analysis, rely on simplified convex geometries, and lack robust real-world implementation. The limitation of convex obstacle representations is specifically addressed in \cite{ecker:iros:2025}, which proposes an efficient Euclidean distance field (EDF)-based collision detection routine for global planning in environments obtained from LiDAR data.


Finally, for completeness, we note that collision-free MPC is a widely adopted strategy in other robotic domains, such as legged robots~\cite{gaertner:2021,chiu:2022}, aerial vehicles~\cite{lindqvist:2020} and mobile autonomous ground vehicles~\cite{pankert:2020}. However, these applications typically involve systems with fast actuation and low inertia, unlike the large-scale underactuated dynamics and hydraulic constraints of forestry cranes. 

\subsection{Contribution}
This work presents the first collision-free sway-damping MPC for forestry cranes that directly integrates LiDAR-based perception into the control loop with experimental validation on real hardware. Prior approaches treat sway damping and collision avoidance as separate problems: existing controllers either focus on sway damping while assuming predefined collision-free paths \cite{kalmari:2014,kalmari:2017}, or handle collision avoidance exclusively at the global planning level without considering payload dynamics \cite{ayoub:2024,spinelli:2025}. Our unified MPC framework addresses both challenges simultaneously, enabling truly reactive behavior that adapts to environmental changes while maintaining payload stability.
This integration provides three key advantages: (i) real-time navigation around unforeseen quasi-static environmental changes, (ii) reduced dependency on precise trajectory tracking through embedded collision constraints, and (iii) environment-aware disturbance rejection that maintains safety under external perturbations.
Building on previous theoretical work \cite{ecker:2025,ecker:iros:2025}, this paper contributes:
\begin{itemize}
\item Experimental demonstration of reactive behavior under unforeseen environmental changes and external disturbances,
\item Real-time integration of online EDF mapping into the MPC framework, eliminating reliance on predefined convex obstacle representations,
\item First experimental validation of unified collision-free sway-damping control on an operational forestry crane.
\end{itemize}
As a proof of concept, this work establishes the feasibility of unifying sway damping and collision avoidance in a single MPC framework for real forestry operations, providing the foundation for future autonomous forestry applications.


\section{Forestry Crane Model}
We leverage the crane model presented in the previous work \cite{ecker:2025,ecker:ifac:2025}. We split the joint space into actuated and passive joints (pendulum dynamics), i.e. $\mathbf{q}^\mathrm{T}=[\mathbf{q}_A^{\mathrm{T}},\mathbf{q}_P^\mathrm{T}]\in\mathds{R}^7$ with $\mathbf{q}_A\in\mathds{R}^5$ and $\mathbf{q}_P\in\mathds{R}^2$. We reasonably assume that the opening angle of the jaw remains constant during point-to-point planning. This assumption reduces the gripper to a single rigid body.

\subsection{Pendulum Dynamic Model}

 We can express the pendulum dynamics as
 \begin{align}\label{eq:PendulumDynamics}
    \ddot{\mathbf{q}}_P&=-\mathbf{D}_{P}^{-1}(\mathbf{q})(\mathbf{D}_{M}(\mathbf{q})\ddot{\mathbf{q}}_A+\mathbf{C}_P(\mathbf{q},\dot{\mathbf{q}})\dot{\mathbf{q}}+\mathbf{g}_P(\mathbf{q})) \Comma
\end{align}
where  $\mathbf{D}_{M}(\mathbf{q})\in\mathds{R}^{2\times 5}$ and $\mathbf{D}_{P}(\mathbf{q})\in\mathds{R}^{2\times 2}$ are the corresponding blocks of the mass matrix, $\mathbf{C}_P\in\mathds{R}^{2\times 7}$ of the Coriolis matrix, and the potential forces $\mathbf{g}_P(\mathbf{q})\in\mathbb{R}^{2}$.

\subsection{Hydraulic Actuator Model}
\begin{figure}
\centering
\includegraphics[trim=0cm 0.6cm 0cm 0cm,clip,scale=0.6]{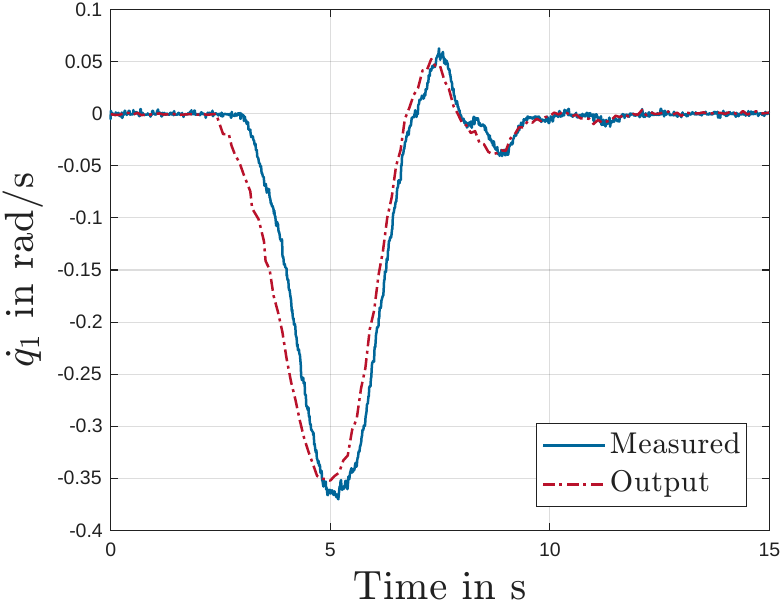}
\includegraphics[trim=0cm 0cm 0cm 0cm,clip,scale=0.6]{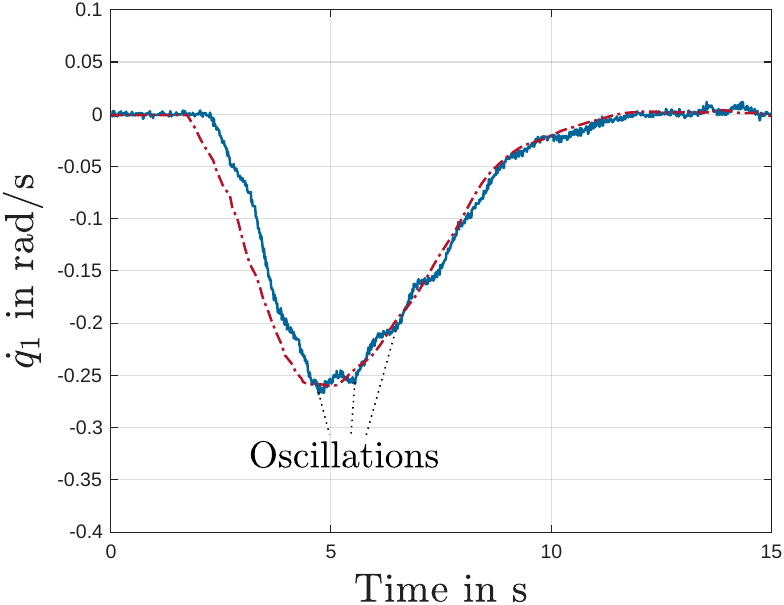}
\caption{Comparison of the closed-loop behavior using first and second order actuator models. Blue: Measured closed-loop joint velocity. Red dashed: Controller output $u$. Top: Slewing velocity with second order model. Bottom: Slewing velocity with first order model.}\label{fig:ActuatorModel}
\end{figure}
For the slewing joint, we observed that direct velocity outputs \cite{ecker:2025} resulted in instabilities, making them unsuitable for closed-loop control. In contrast, first-order actuator models \cite{kalmari:2014,kalmari:2017} produced stable behavior, although oscillations in the closed-loop joint velocities could still be observed, as shown in the lower part of Fig.~\ref{fig:ActuatorModel}. Motivated by the approach in \cite{nan:2024}, we therefore adopt the following second-order velocity model to describe the actuator dynamics
\begin{align}\label{eq:ActuatorDynamics}
\dddot{q}_{A} = \omega^2(u-\dot{q}_{A}) - 2d\omega \ddot{q}_{A} \Comma
\end{align}
where $\omega$ is the natural frequency and $d$ is the damping constant. Steady-state nonlinearities are handled via lookup tables that map the controls $u$ to the joystick commands, following \cite{spinelli:2025}.  
This model captures the mapping between the control inputs $\mathbf{u}^{\mathrm{T}} = [u_1, \dots, u_{5}]$ and the actuator velocities $\dot{\mathbf{q}}_A$. As demonstrated in the top part of Fig.~\ref{fig:ActuatorModel}, the second-order formulation significantly reduces oscillations in the closed-loop joint velocities compared to conventional first-order models \cite{kalmari:2014,kalmari:2017}.

\subsection{Pump flow rate}\label{sec:Total pump flow rate}
The cylinders are supplied by a single pump, resulting in a pump flow rate for the hydraulic system given by
\begin{align}\label{eq:TotalPumpFlow}
    Q(t) &= \sum_{l=1}^{5}A_l\big(\mathrm{sign}(\dot{d}_l)\big)|\dot{d}_l| \ ,
\end{align}
where $A_l(\cdot)$, $l=1,\dots,5$ are the direction-dependent effective areas of the cylinders and $d_l$, $l=1,\dots,5$ are the cylinder displacements. The pump flow rate (\ref{eq:TotalPumpFlow}) that can be supplied by the hydraulic actuation system is limited, yielding the pump flow rate constraint (PFRC) $Q(t)\leq Q_{\mathrm{max}}$, c.f. \cite{ecker:2025}.

\section{Collision Free Sway Damping MPC for Safe Forestry Crane Navigation}
Fig.~\ref{fig:Schematic} illustrates the overall navigation pipeline for the forestry crane, consisting of 3 different modules:
\begin{enumerate}
    \item a perception module, for environment mapping,
    \item a global planning module, for computing a global reference connecting the actual configuration to the target endeffector (EE) pose,
    \item a local planning module, which implements our collision-free MPC for reactive behavior.
\end{enumerate}
The collision-free MPC, which is the main contribution of this work, receives reference trajectories $\mathbf{q}_{A,d}(t)$ for the actuated joints from the global planner \cite{ecker:2025,ecker:iros:2025}, the EDF map from the perception module \cite{hornung:2013,han:2019} as well as the current joint positions and velocities.

\subsection{Perception}
The perception module processes the crane's joint measurements along with point cloud data \(\mathcal{P}\) from the LiDAR. Points corresponding to the crane itself are removed using a point cloud filter. An occupancy map is constructed using the OctoMap framework \cite{hornung:2013}, while the voxelized EDF is incrementally updated via the FIESTA method \cite{han:2019}. The updated EDF is then provided to the MPC. All necessary coordinate transformations are computed from the joint angle measurements \(\mathbf{q}\).

\subsection{Global Planning}
We employ the PFRC-VP-STO global planner from \cite{ecker:2025}, combined with EDF-based collision detection \cite{ecker:iros:2025}. The global planner generates a reference trajectory $\mathbf{q}_{A,d}(t)$ in form of a cubic spline for the MPC, while ignoring the dynamics of the passive joints. The target configuration is determined via inverse kinematics (IK), which computes the desired joint values from a user defined 4D EE-pose $(x,y,z,\varphi)$, where $x,y,z$ are the position coordinates and $\varphi$ is the yaw of the gripper.

\subsection{MPC Formulation}
The model predictive control (MPC) problem is posed as the following optimal control problem:
\begin{subequations}\label{eq:DiscreteTimeOCPProblem}
\begin{align}
    \min\limits_{\mathbf{X},\mathbf{U},\boldsymbol{\tau},\dot{\boldsymbol{\tau}}} &\sum_{k=k_0}^{k_0+N-1} l(\mathbf{x}_k,\mathbf{u}_k, \tau_k, \dot{\tau}_k)\label{eq:DiscreteTimeOCPProblem objective}    \\
    \text{s.t.}\;  &\mathbf{x}_{k+1}=\mathbf{f}(\mathbf{x}_k,\mathbf{u}_k) \ ,\; \mathbf{x}_{k_0}=\mathbf{x}_{\mathrm{init},k_0}\label{eq:OCPDynamics}\\
    &\tau_{k+1}=\tau_k+\dot{\tau}_kT_s
    \\
    & \mathbf{q}_k\in[\mathbf{q}_{\mathrm{min}},\mathbf{q}_{\mathrm{max}}]\ ,\; Q\leq Q_{\mathrm{max}}\label{eq:OCPlimitsOCPvolumeFlowConstaints}
    \\
    & \ddot{\mathbf{q}}_{A,k}\in[\ddot{\mathbf{q}}_{A,\mathrm{min}},\ddot{\mathbf{q}}_{A,\mathrm{max}}]\label{eq:OCPAcclLimits}\\
    &    \mathrm{sd}_{i}(\mathbf{q})> 0 \ , i\in\{1,2,3\}    \label{eq:OCPInequalities} \ \Comma
\end{align}
\end{subequations}  
where $\mathbf{X} = [\mathbf{x}_{k_0},\dots,\mathbf{x}_{k_0+N-1}]$ collects the state vectors $\mathbf{x}^\mathrm{T} = [\mathbf{q}^\mathrm{T}, \dot{\mathbf{q}}^\mathrm{T}, \ddot{\mathbf{q}}_{A}^\mathrm{T}]$, $\mathbf{U} = [\mathbf{u}_{k_0},\dots,\mathbf{u}_{k_0+N-1}]$ contains the control inputs, and $\boldsymbol{\tau} = [\tau_{k_0},\dots,\tau_{k_0+N-1}]^\mathrm{T}$ together with $\dot{\boldsymbol{\tau}} = [\dot{\tau}_{k_0},\dots,\dot{\tau}_{k_0+N-1}]^\mathrm{T}$ denote the time progress variable and its rate, respectively. The discrete-time dynamics \eqref{eq:OCPDynamics} contains the actuator model \eqref{eq:ActuatorDynamics} and the pendulum dynamics \eqref{eq:PendulumDynamics}.
The main motivation of including the time progress variable as an additional optimization variable is to enable robust stopping in case the MPC is not able to find a solution around the obstacle. 

The constraints \eqref{eq:OCPlimitsOCPvolumeFlowConstaints},\eqref{eq:OCPAcclLimits} and \eqref{eq:OCPInequalities} enforce joint and acceleration limits, bound the hydraulic pump flow rate by $Q_{\mathrm{max}}$, and ensure collision avoidance via signed distance conditions.
Constraint violations are penalized using a quadratic barrier function with margin $\varepsilon$ and penalty weight $\mu = \frac{10}{\varepsilon}$, following the formulation in \cite{gaertner:2021}. The dependence of $\mu$ and $\varepsilon$ ensures a constant penalty per time step violations.

\subsubsection{Costs}
The stage cost $l(\cdot)$ consists of five components
\begin{align}
\begin{aligned}
    l(\mathbf{x},\mathbf{u},\tau,\dot{\tau}) &= l_{\mathrm{track}}(\mathbf{q}_A,\tau) +  l_{\mathrm{damp}}(\dot{\mathbf{q}}_P)
    +l_{\mathrm{vel}}(\dot{\mathbf{q}}_A)\\&+l_{\mathrm{accl}}(\ddot{\mathbf{q}}_A) + l_{\mathrm{prog}}(\dot{\tau})\FullStop
\end{aligned}
\end{align} 

The tracking and progress costs are given by
\begin{align}
\begin{aligned}
    l_{\mathrm{track}}(\mathbf{q}_A,\tau) &= w_{\mathrm{track}}\|\mathbf{q}_{A}-\mathbf{q}_{A,d}(\tau)\|^2\\
    l_{\mathrm{prog}}(\dot{\tau})&=w_{\mathrm{prog}}\|\dot{\tau}-1\|^2
    \Comma
\end{aligned}
\end{align} 
where $\mathbf{q}_{A,d}(\cdot)$ is the reference trajectory for the actuated joints from the global planner.
To suppress passive joint oscillations, a damping cost is introduced
\begin{align}
    l_{\mathrm{damp}}(\dot{\mathbf{q}}_P) &= w_{\mathrm{damp}}\|\dot{\mathbf{q}}_P\|^2\FullStop
\end{align}  

Finally, smoothness is promoted by penalizing joint velocities and accelerations
\begin{align}
    l_{\mathrm{vel}}(\dot{\mathbf{q}}_A) &= w_{\mathrm{vel}}\|\dot{\mathbf{q}}_A\|^2\Comma &
    l_{\mathrm{accl}}(\ddot{\mathbf{q}}_A) &= w_{\mathrm{accl}}\|\ddot{\mathbf{q}}_A\|^2\FullStop
\end{align} 

\subsubsection{Collision Constraints}
To incorporate the voxelized EDF map into the collision-avoidance constraints, we model the boom $\mathcal{L}_1$, the arm $\mathcal{L}_2$, and the gripper $\mathcal{L}_3$ as the links relevant for collision checking \cite{ecker:iros:2025}. Each link $\mathcal{L}_i$, $i=1,\dots,3$, is approximated by a set of spheres $\mathcal{S}_i=\{(\mathbf{p}_{ij}, r_{ij})\}_{j=1}^{M_i}$. To account for the variable length of the telescopic arm, the number of spheres is adjusted to maintain a desired separation along the center line.
For constraint formulation, we select the sphere with the minimum EDF value for each link. The signed distance for link $\mathcal{L}_i$ is then defined as
\[
\mathrm{sd}_{i}(\mathbf{q}) = \min_{j}\{ d_{\mathrm{EDF}}(\mathbf{p}_{ij})-r_{ij}\}
\]
which ensures a conservative approximation of the closest obstacle. This yields three inequality constraints \eqref{eq:OCPInequalities}.

\section{Experimental Results}

The proposed approach was validated on a real forestry crane equipped with a Livox Avia LiDAR. The perception and the global planning module are implemented as standard ROS 2 Nodes on the same IPC. The MPC is implemented as a \texttt{ros2\_control} controller on a separate IPC and runs at a frequency of 10Hz.

We conduct a series of experiments to demonstrate the sway damping and collision avoidance capabilities of our approach:
\begin{enumerate}
\item We first apply disturbances to the gripper in free space to show that the MPC effectively damps the induced motions compared to the uncontrolled case. We then repeat the experiment near an obstacle to demonstrate that the controller simultaneously accounts for collision avoidance while damping sway.
\item We evaluate the system’s ability to adapt to unforeseen environmental changes that are not captured by the global planner.
\item We demonstrate that, even in static environments, the collision-free MPC compensates for tracking inaccuracies to ensure collision avoidance.
\end{enumerate}

The videos demonstrating these experiments can be found at the \texttt{attached video}\footnote{\url{https://youtu.be/tEXDoeLLTxA}}.

\begin{figure*}
\centering
\includegraphics[trim=0cm 0cm 0cm 0cm,clip,scale=0.44]{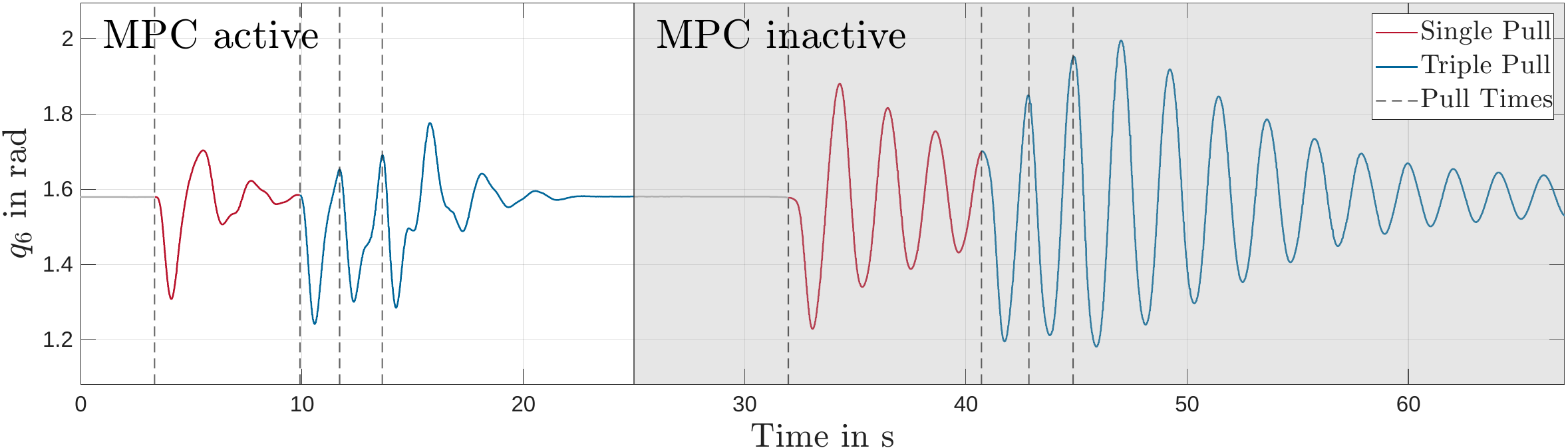}
\caption{Motions of the passive joint $q_6$ under external force disturbance. Left: MPC is active. Right: MPC is inactive. Red: Motions after single pull. Blue: Motions after triple pull.}\label{fig:SwayDamping}
\end{figure*}

\begin{figure}
\centering
\includegraphics[scale=0.66]{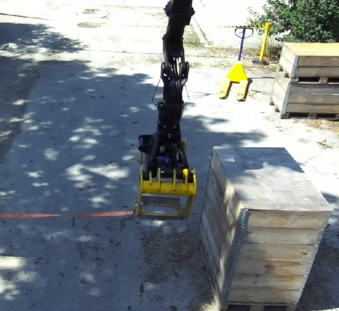}
\caption{Setup for the sway damping experiment.}\label{fig:SwayDampingExperiment}
\end{figure}

\begin{figure}
\centering
\includegraphics[trim=0cm 0cm 0cm 0cm,clip,scale=0.6]{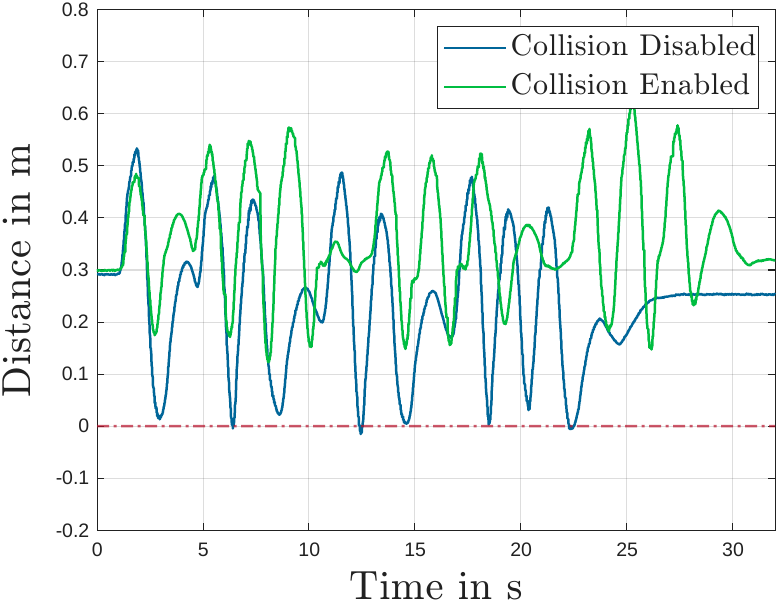}
\caption{Distances for the sway damping experiment. Blue: MPC with collision costs being disabled. Green: MPC with enabled collision costs}\label{fig:SwayDampingCollAvoid}
\end{figure}
\subsection{Parameter tuning}
For cost tuning, $w_{\mathrm{track}}$ was fixed to 1, and the remaining weights were adjusted relative to this value. For pendulum damping, $w_{\mathrm{damp}}=0.1$ provided the best sway attenuation, as larger values resulted in overly sensitive behavior. The acceleration weight was set to $w_{\mathrm{accl}}=0.1$, since smaller values occasionally led to excessively aggressive motions. Finally, $w_{\mathrm{prog}}=0.2$ provided a good trade-off between maintaining fast progress and ensuring safe stopping.

The MPC employs a fixed-time termination criterion of 70ms, ensuring predictable runtime performance suitable for real-time control. Across all experiments, feasible solutions were consistently obtained within this budget, allowing reliable operation at 10 Hz. We did not observe any benefits by further increasing the MPC frequency. This design choice emphasizes robustness over solver optimality, which is particularly important for real-world deployment.
For the prediction horizon length, values of $N$ between $30$ (corresponding to 3s, with the pendulum period at approximately 2s) and $80$ (8s) were evaluated. The MPC remained stable for all tested values. A horizon length of $N=40$ (4s) was selected, as it exhibited improved convergence when bypassing obstacles compared to shorter horizons, while no further benefits were observed for larger values.

Finally,  we observed that a minimum penalty safety margin of $\varepsilon=0.2$m was possible to obtain robust collision avoidance. For smaller values the cost increase becomes too steep, yielding a too aggressive behavior of the MPC which results in instabilities close to obstacles.



\subsection{Sway damping}
To evaluate the sway damping performance, external disturbances are introduced to the passive joints using a rope, as shown in Figure~\ref{fig:SwayDampingExperiment}.
First we evaluate the performance in free space, i.e. without any obstacles nearby the gripper. The experiment consists of two scenarios: a single pull and a sequence of three pulls at the system’s resonance frequency. The resulting motion of the passive joint $q_6$ is visualized in Fig.~\ref{fig:SwayDamping}. The left part shows the joint angle with the MPC activated, while the right part presents the behavior without MPC. While the gripper stops swinging after $\approx$5s with activated MPC, it oscillates more than 20s without MPC. This result clearly demonstrates that the MPC effectively dampens sway in the passive joints.

In the second step of the experiment, the gripper is positioned in close proximity to an obstacle, as depicted in Fig.~\ref{fig:SwayDampingExperiment}. The corresponding distance profiles under external force disturbances are shown in Fig.
~\ref{fig:SwayDampingCollAvoid}. When collision costs are disabled in the MPC (blue), the gripper makes contact with the box. In contrast, enabling collision costs (green) effectively prevents collisions and ensures a consistent safety margin to the obstacle. These results highlight the capability of the MPC to incorporate environmental information from LiDAR feedback, thereby maintaining safe operation even in the presence of external disturbances. A demonstration of these experiments is provided in the \texttt{attached video}.
\begin{figure*}[t]
\centering
\includegraphics[scale=0.34]{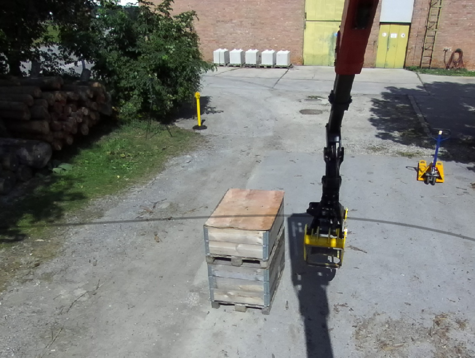}
\includegraphics[scale=0.34]{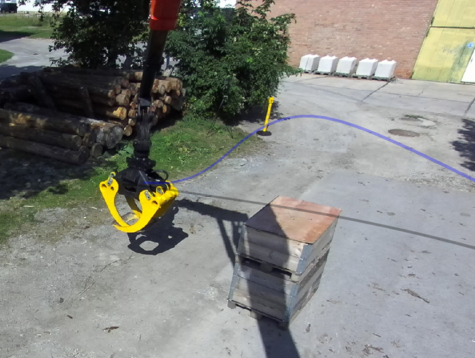}
\includegraphics[scale=0.34]{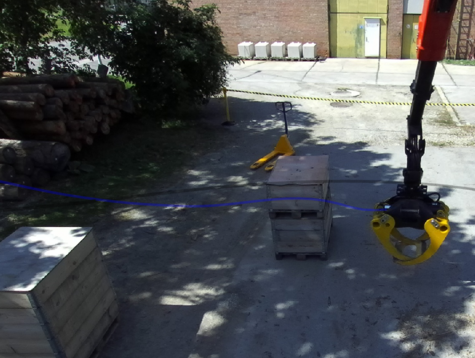}
\includegraphics[scale=0.34]{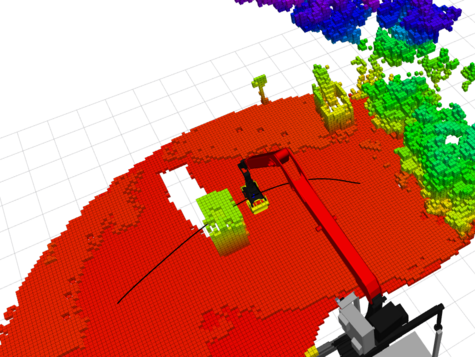}
\includegraphics[scale=0.34]{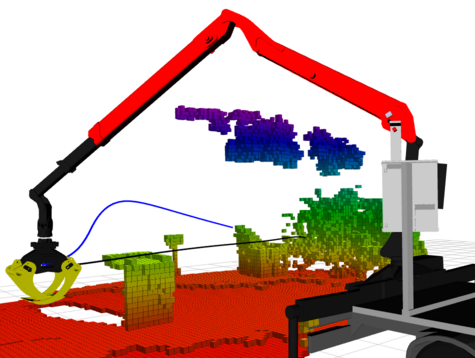}
\includegraphics[scale=0.34]{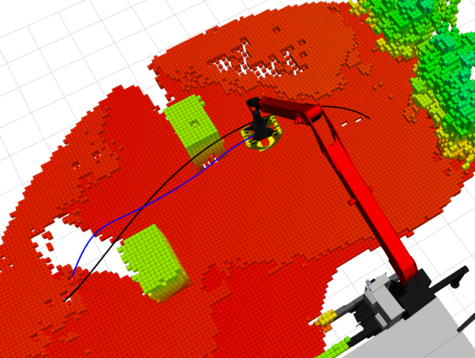}
\caption{MPC Replanning Experiments. The black lines illustrate the reference trajectory of the global planner and the blue lines the MPC horizon in task-space Left: MPC stops in front of obstacle. Middle: MPC has to avoid one additional obstacle. Right: MPC has to avoid two additional obstacles.}\label{fig:ReplanningTest}
\end{figure*}
\subsection{Replanning}
\begin{figure*}[t]
\centering
\includegraphics[scale=0.34]{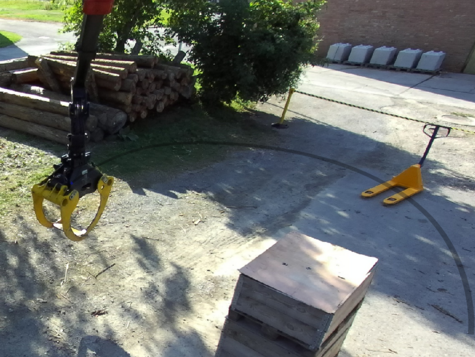}
\includegraphics[scale=0.34]{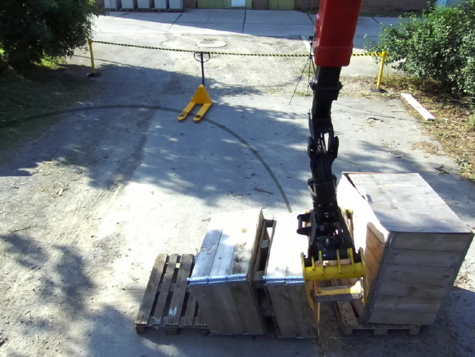}
\includegraphics[scale=0.34]{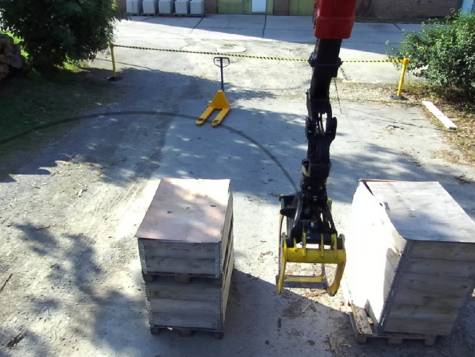}
\includegraphics[scale=0.34]{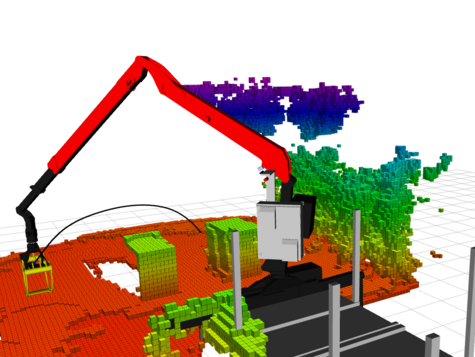}
\includegraphics[scale=0.34]{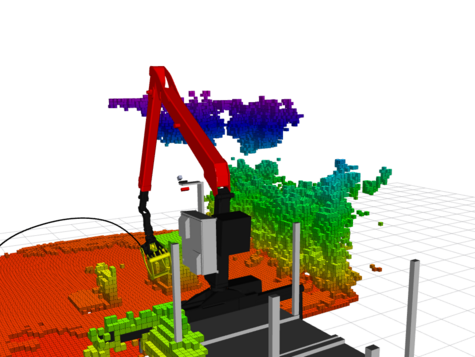}
\includegraphics[scale=0.34]{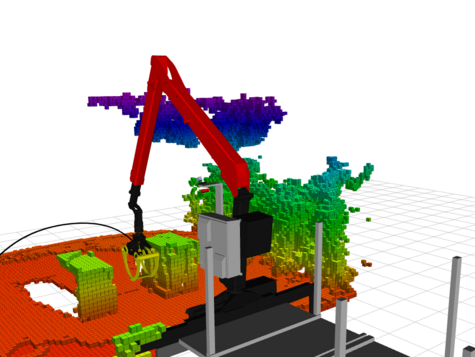}
\caption{Narrow target experiment. Left: Control task illustration. Middle: Result with blind sway damping MPC. Right: Result with collision free sway damping MPC.}\label{fig:SlipTask}
\end{figure*}

We demonstrate the replanning capabilities of the MPC for different scenarios, three of them given in Fig.~\ref{fig:ReplanningTest}. The global reference trajectory in task-space is visualized by the black lines, whereas the local MPC horizon is shown in blue. In all cases, we place boxes s.t. they collide with the reference of the global planner.

\textbf{Stopping:} In practical applications, situations may arise in which the MPC is unable to generate a feasible trajectory to bypass an obstacle. This limitation may result from the absence of a valid local solution or from insufficient global planning capabilities. To guarantee safe navigation in such cases, the MPC must be capable of coming to a complete stop in front of the obstacle, thereby preventing potential damage. The first scenario illustrates this behavior: a box is placed such that the MPC cannot compute a collision-free trajectory and is therefore required to stop. The black line in the left image of Fig.~\ref{fig:ReplanningTest} represents the global planner’s reference trajectory in task space, which intersects with the obstacle. In contrast, the MPC halts, with all predicted trajectory points within the horizon converging to approximately the same position. As a result, the blue line is not visible.

\textbf{Bypassing a single obstacle:} In the second scenario, the start and goal configurations are modified such that the gripper is positioned slightly higher, enabling the MPC to compute a feasible bypass trajectory. The global reference trajectory in task space, shown as the black line in the middle image of Fig.~\ref{fig:ReplanningTest}, intersects with the obstacle. However, the MPC trajectory, illustrated by the blue line in the same image, deviates from the reference path and successfully guides the crane around the obstacle.
    
\textbf{Bypassing multiple obstacles:} In the third scenario, the setup is extended by introducing two additional obstacles that also intersect the global reference trajectory (black line in Fig.~\ref{fig:ReplanningTest}). The MPC, however, adapts its local trajectory within the prediction horizon, as illustrated by the blue line, and successfully replans a collision-free trajectory. This enables the crane to reach the goal configuration without colliding with any of the obstacles.

In summary, our experiments demonstrate the replanning capabilities of our collision-free MPC, thus ensuring safe operation in quasi-static environments. The controller is able to bypass obstacles not present during the global planning stage and stops if it is not able to find a collision free bypass trajectory. We again refer to the \texttt{attached video} for demonstrations of these experiments.

\subsection{Tracking Inaccuracies}
\begin{figure}
\centering
\includegraphics[trim=0cm 0.6cm 0cm 0cm,clip,scale=0.6]{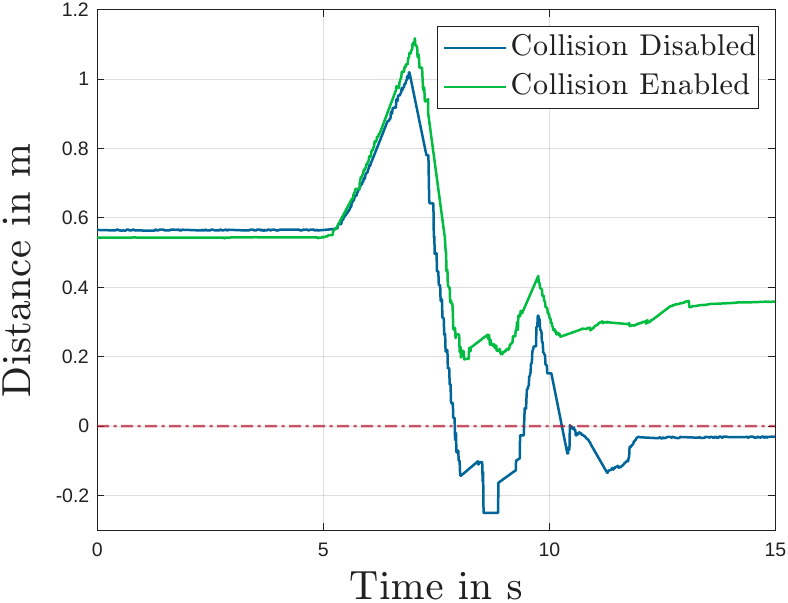}
\includegraphics[trim=0cm 0cm 0cm 0cm,clip,scale=0.6]{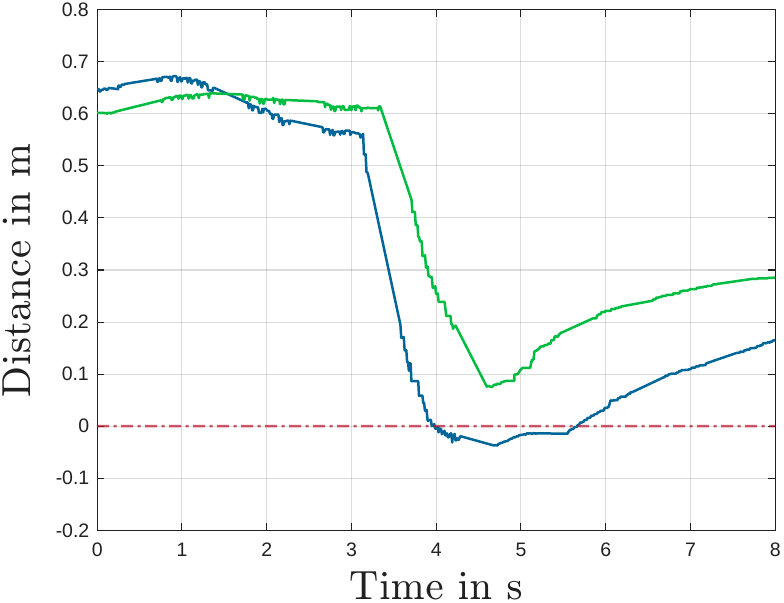}
\caption{EDF distances for the narrow gap and close obstacle stop experiments. Blue: Collision constraints disabled, Green: Collision constraints enabled. Red dashed: Zero distance limit, i.e. collision. Top: Narrow gap experiment. Bottom: Close obstacle stop experiment}\label{fig:DistancesGap}
\end{figure}


Finally, the controller was evaluated on two tasks in which the global planner had full knowledge of the environment and the crane was required to operate in close proximity to obstacles: (i) moving through a narrow gap and (ii) stopping near an obstacle.

The left image of Fig.~\ref{fig:SlipTask} illustrates the first task, where the crane had to stop within a narrow passage between two obstacles.
%
The top part of Fig.~\ref{fig:DistancesGap} shows the minimum gripper distances for the narrow-gap experiment. With collision constraints deactivated (blue), the controller was unable to reach the goal without collisions, as also illustrated in the middle image of Fig.~\ref{fig:SlipTask}. When the collision constraints were activated (green), the controller successfully guided the crane to the goal without collisions and without any additional parameter modifications, as shown in the right image of Fig.~\ref{fig:SlipTask}.

The second experiment required the crane to follow a trajectory that stopped close to an obstacle. The minimum gripper distances for this experiment are shown in the bottom part of Fig.~\ref{fig:DistancesGap}. Deactivating the collision constraints again led to a collision (blue), whereas activating them enabled a collision-free stop (green).

These experiments demonstrate that safe operation can be ensured without relying on highly precise trajectory tracking, which is particularly challenging for such machines. Video recordings of both experiments are provided in the \texttt{attached video}.

\subsection{Limitations}\label{sec:Limitations}
This section outlines the limitations of the proposed approach. Since the present work primarily serves as a proof of concept for a collision-free sway-damping MPC, these limitations do not invalidate the results but highlight directions for future research.

\textbf{Quasi-Static Environments:} The current approach is restricted to quasi-static environmental changes, as this assumption underlies state-of-the-art EDF mapping methods \cite{han:2019,oleynikova:2017,pan:2022}. To enable operation in fully dynamic environments, extensions such as \cite{schmid:2023} are required. Furthermore, an additional safety mechanism, such as directly evaluating the MPC horizon on the point cloud, could be beneficial to trigger an emergency stop if the mapping process is too slow.

\textbf{Limited Sensor FoV:} The employed LiDAR sensor provides a limited field of view, resulting in large unobserved regions behind obstacles. Consequently, the planner may generate seemingly collision-free trajectories that pass through such unseen areas. This limitation could be mitigated through additional sensor mountings or by adapting the mapping methodology to better account for occlusions.

\textbf{Global Replanning:} While the experiments demonstrate that the MPC safely halts in front of obstacles when no bypass solution is available, real-world applications require continued task execution in such scenarios. This necessitates reliable detection of dead-end situations and subsequent re-invocation of the global planner to resume operation.

\textbf{Full Log Loading Task:} The controller demonstrated stable behavior when handling logs with known dimensions and mass properties. However, integration with the perception pipeline introduces additional challenges, including reliable detection and accurate volume estimation of the log. Furthermore, during the grasping process, the log transitions from being an external object to becoming part of the crane’s collision model. This transition must be properly managed by the components in the perception module, such as the point cloud filter.

\section{Conclusion \& Future Work}
This work introduces a collision-free sway damping model predictive controller for safe navigation of a forestry crane. We integrate the controller in a navigation pipeline, including global planning and perception for environment mapping. Our experiments demonstrate that the controller dampens pendulum sway, while accounting for environmental sensor feedback from the LiDAR. This enables reactive behavior due to unforeseen quasi-static changes in the environment, inaccurate tracking control as well as in case of external disturbances. By demonstrating feasibility on real hardware, this proof of concept provides the foundation for subsequent work on full log-handling tasks, while addressing the limitations from Section~\ref{sec:Limitations}.

\bibliographystyle{plain} 
\bibliography{refs} 

\end{document}